\documentclass[letterpaper]{article} 
\usepackage{aaai2026}  
\usepackage{times}  
\usepackage{helvet}  
\usepackage{courier}  
\usepackage[hyphens]{url}  
\usepackage{graphicx} 
\usepackage{bm}
\usepackage{natbib}  
\usepackage{caption} 
\usepackage{algorithm}
\usepackage{algorithmic}
\usepackage{multirow}
\usepackage{booktabs}
\usepackage{amssymb}
\usepackage{bbding}
\usepackage[table]{xcolor}
\definecolor{first}{RGB}{191, 225, 201} 
\definecolor{second}{RGB}{227, 237, 185} 
\definecolor{third}{RGB}{254, 250, 194} 

\usepackage{amsmath}

\usepackage{newfloat}
\usepackage{listings}
\DeclareCaptionStyle{ruled}{labelfont=normalfont,labelsep=colon,strut=off} 
\floatstyle{ruled}
\newfloat{listing}{tb}{lst}{}
\floatname{listing}{Listing}
\title{DGSAN: Dual-Graph Spatiotemporal Attention Network for 
Pulmonary Nodule Malignancy Prediction}
\author{
    Xiao Yu\textsuperscript{\rm 1}, Zhaojie Fang\textsuperscript{\rm 1,2}, Guanyu Zhou\textsuperscript{\rm 1}, Yin Shen\textsuperscript{\rm 1}, 
    Huoling Luo\textsuperscript{\rm 3},\\
    Ye Li\textsuperscript{\rm 4,5},
    Ahmed Elazab\textsuperscript{\rm 6},
    Xiang Wan\textsuperscript{\rm 7},
    Ruiquan Ge\textsuperscript{\rm 1,5},
    Changmiao Wang\textsuperscript{\rm 7,}\thanks{Corresponding author: Changmiao Wang}
}
\affiliations{
    \textsuperscript{\rm 1}Hangzhou Dianzi University, Hangzhou, China\\
    \textsuperscript{\rm 2}The Chinese University of Hong Kong, Shenzhen, Shenzhen, China\\
    \textsuperscript{\rm 3}Shenzhen Institute of Information Technology, Shenzhen, China\\
    \textsuperscript{\rm 4} Shenzhen Institutes of Advanced Technology, Chinese Academy of Sciences\\
    \textsuperscript{\rm 5}Hangzhou Institute of Advanced Technology, Hangzhou, China\\
    \textsuperscript{\rm 6}Tsinghua Shenzhen International Graduate School, Tsinghua University, Shenzhen, China\\
    \textsuperscript{\rm 7}Shenzhen Research Institute of Big Data, Shenzhen, China\\

    Corresponding author*: cmwangalbert@gmail.com.

}

\begin{document}

\maketitle

\begin{abstract}
Lung cancer continues to be the leading cause of cancer-related deaths globally. Early detection and diagnosis of pulmonary nodules are essential for improving patient survival rates. Although previous research has integrated multimodal and multi-temporal information, outperforming single modality and single time point, the fusion methods are limited to inefficient vector concatenation and simple mutual attention, highlighting the need for more effective multimodal information fusion. To address these challenges, we introduce a Dual-Graph Spatiotemporal Attention Network, which leverages temporal variations and multimodal data to enhance the accuracy of predictions. Our methodology involves developing a Global-Local Feature Encoder to better capture the local, global, and fused characteristics of pulmonary nodules. Additionally, a Dual-Graph Construction method organizes multimodal features into inter-modal and intra-modal graphs. Furthermore, a Hierarchical Cross-Modal Graph Fusion Module is introduced to refine feature integration. We also compiled a novel multimodal dataset named the NLST-cmst dataset as a comprehensive source of support for related research. Our extensive experiments, conducted on both the NLST-cmst and curated CSTL-derived datasets, demonstrate that our DGSAN significantly outperforms state-of-the-art methods in classifying pulmonary nodules with exceptional computational efficiency. 
\end{abstract}

\begin{links}
    \link{Code}{https://github.com/lcbkmm/DGSAN}
\end{links}

\section{Introduction}
\label{sec:intro}
Lung cancer is the leading cause of cancer-related mortality globally, with approximately 2.5 million new cases diagnosed annually, accounting for 12.4\% of all newly diagnosed cancers. It is responsible for around 1.8 million deaths, which represent 18.7\% of the total cancer mortality rate~\cite{Bray2024GlobalCS}. Unfortunately, lung cancer has a poor prognosis, with a low five-year survival rate. The survival rate for early-stage patients is approximately 50\%–60\%, while for those diagnosed at advanced stages, it drops to less than 10\%~\cite{henschke202320}. In clinical practice, initial low-dose computed tomography (CT) screening often detects a significant number of lung nodules with diameters ranging from 5–10 mm. However, a single imaging scan is insufficient to differentiate between benign proliferations, inflammatory nodules, and early malignant lesions, requiring multiple follow-up scans to monitor their dynamic progression. Relying solely on static features such as nodule size, shape, and CT values can lead to missed diagnoses and misinterpretation of small or atypical lesions. Therefore, the early detection and dynamic monitoring of lung nodules are critical for improving diagnostic accuracy and enhancing the survival rates of lung cancer patients.

In recent years, although some studies have demonstrated that single-image modalities can achieve good results \cite{10980792} in classification and prediction, the lack of spatio-temporal and modality transformation often makes the analysis of practical problems insufficiently comprehensive. Now, the approach to predicting the malignancy of pulmonary nodules has progressed from static single-modality analysis to dynamic multi-modality fusion. Jiang \textit{et al.}~\cite{jiang2021learning} introduced an attention mechanism for single-time-point CT data, but this method did not effectively capture the dynamic changes in lesions. As interest in lesion dynamics increased, research shifted towards analyzing data across multiple time points. Liu \textit{et al.}~\cite{liu2022study} developed a time-modulated LSTM network incorporating a time interval weighting mechanism, which better captures the spatiotemporal growth patterns of lesions. Building on this, Song \textit{et al.}~\cite{song2023msts} tackled the challenge of modeling the heterogeneous growth rates of nodules with a dynamic adaptive time module.

\begin{figure*}[t]
    \centering
    \includegraphics[width=1\textwidth]{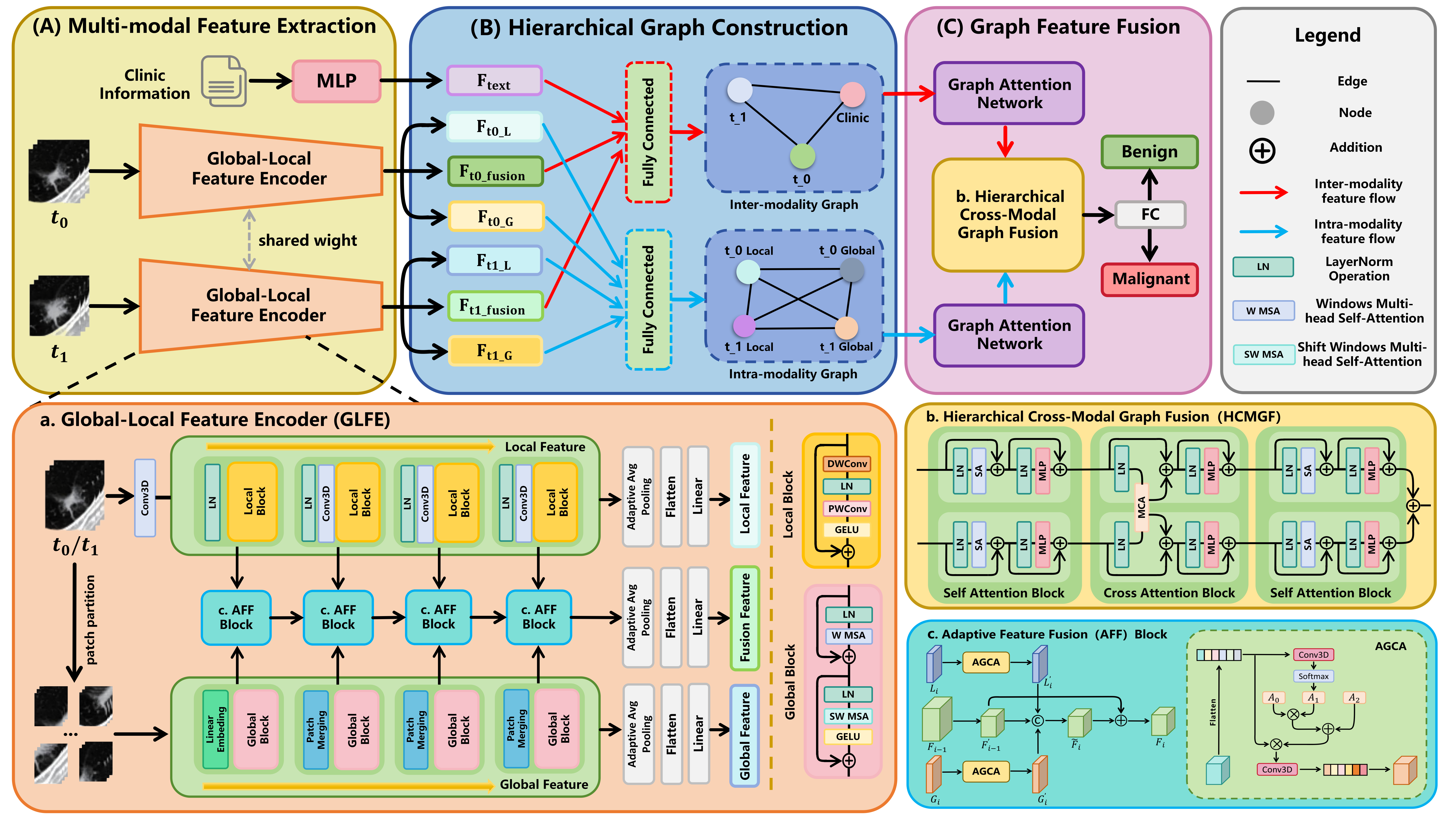}
    \caption{Overall framework of the proposed DGSAN, which comprises three parts: (A) Multi-modal Feature Extraction, designed to fuse local and global information and generate high-dimensional spatiotemporal representations, (B) Hierarchical Graph Construction, builds intra- and inter-modal graphs based on imaging and clinical features, using graph attention to model complex spatial–temporal and cross-modal dependencies, and (C) Graph Feature Fusion, utilizes a “self → cross → self” attention mechanism to align multimodal information and produce unified representations for malignancy prediction.}
    \label{fig1}
\end{figure*}

The focus of current research has transitioned from single-modality to multi-modality spatiotemporal fusion. Yu \textit{et al.}~\cite{yu2024ichpro} designed a joint attention mechanism to integrate CT images with clinical text, while Shen \textit{et al.}~\cite{10981213} proposed a cross-modality attention mechanism to align follow-up CT features bidirectionally with clinical data. Despite advancements, multi-time series analysis struggles with tracking dynamic lesion changes over time, and multimodal fusion faces challenges in reconciling differences between various modalities.

Traditional convolutional methods, primarily intended for data in Euclidean space, fall short when addressing intricate relationships and interdependencies among objects. To effectively capture complex interactions and higher-order dependencies in multimodal data, Graph Convolutional Networks (GCNs) and their variants have shown remarkable potential in improving multimodal fusion. Cai \textit{et al.}~\cite{cai2024hierarchical} introduced a hierarchical hypergraph attention network for histopathological survival analysis, achieving synergistic modeling of morphological and topological features. However, this approach suffers from poor adaptability due to manually defined hyperedge rules and incurs high computational costs. Jiang \textit{et al.}~\cite{jiang2024mgdr} proposed a multimodal graph decoupling method using contrastive learning and a multimodal attention module to optimize fusion features. Venkatraman \textit{et al.}~\cite{venkatraman2025sagvitscaleawarehighfidelitypatching} further enhanced image classification performance by combining GCN with Vision Transformers.

Beyond these examples, GCNs have also achieved significant advancements in multimodal cancer survival prediction~\cite{jiang2024hyperrep,su2024cross}, multimodal self-supervised representation learning, and brain disease prediction~\cite{FANG2023102707}. These studies underscore the advantages of GCNs in multimodal fusion, highlighting their adaptability and effectiveness in handling complex data. Moreover, GCNs exhibit a unique and powerful capacity to integrate diverse data types and capture intricate relationships, making them highly valuable in modern multimodal analysis.


To address the challenges associated with integrating diverse types of data collected over various times, we present the Dual-Graph Spatiotemporal Attention Network (DGSAN). This method is designed to explore complex relationships among features both within and across different data modalities. Our key contributions are as follows:

\begin{itemize}

    \item  We developed a Global-Local Feature Encoder (GLFE) that captures voxel-level texture, long-range contextual cues and their fused representations from successive 3D nodule scans, achieving both computational efficiency and superior discriminative power across time points.

    \item We formulated a dual-graph multimodal fusion framework by adopting a graph-based approach that constructs dedicated intra- and inter-modal affinity graphs to explicitly model complex spatial–temporal and cross-modal dependencies, thereby synthesizing image and clinical features more effectively than naïve concatenation.

    \item We implemented a Hierarchical Cross-Modal Graph Fusion Module (HCMGFM) that employs a progressive “self → cross → self” attention architecture to iteratively refine and align high-order semantic information, ensuring robust, noise-resilient integration across dual-modality graphs.

    \item We compiled and publicly released the NLST-cmst dataset, a rigorously annotated longitudinal multimodal cohort of volumetric CT time series and rich clinical variables, thereby establishing a new standard resource to underpin dynamic malignancy prediction research.

\end{itemize}

\section{Methods}
\label{sec:methods}
This study proposes a DGSAN model for predicting the malignancy of pulmonary nodules, as shown in \textbf{Fig.}~\ref{fig1}. First, data from initial ($t_0$) and follow-up ($t_1$) CT scans are input into a shared GLFE to extract local, global, and fused features, while clinical features are obtained via a multilayer perceptron. Next, inter- and intra-modal graphs are constructed from these features, and a Graph Attention Network (GAT)~\cite{liu2020graph} extracts graph features. Finally, these features are further processed and integrated into the HCMGFM, achieving the fusion of dual-modality semantics. The details encompass feature extraction in the GLFE, dual-graph construction, and graph feature fusion within the HCMGFM.

\begin{figure*}[]
    \centering
    \includegraphics[width=0.8\linewidth]{Picture/figure2.png}
    \caption{Illustration of correlations in pulmonary nodule diagnosis. (a) Inter-modal correlations, (b) Intra-modal correlations, (c) A relationship diagram with inter-modal and intra-modal information.
}
    \label{fig2}
\end{figure*}

\subsection{Global-Local Feature Encoder} 
\label{S1}
In order to capture nodule-specific details such as texture and edges (local features) at the voxel level while modeling the long-range dependencies and contextual relationships between the lung nodule and surrounding tissues (global features), we design a Global-Local Feature Encoder. To achieve effective Global-Local Feature, the GLFE employs a three-branch structure. Each branch is dedicated to capturing local features, global features, and fused features, respectively, and consists of four stages. For local feature extraction, the raw image is directly fed into the local feature extraction branch. In this branch, depthwise convolution is first applied to preserve channel information while reducing the computational cost. Following layer normalization, pointwise convolution is used to mix and reorganize the channels of the feature map. This process captures correlations between different channels, thereby enhancing the feature representation. 

In the global feature extraction process, the raw image is divided into 4×4×2 patches, which are passed into the global feature extraction branch. The global feature block uses Windows Multi-head Self-Attention (W-MSA) and Shifted Windows Multi-head Self-Attention (SW-MSA) from the Swin Transformer~\cite{Liu_2021_ICCV}. These modules improve local information interaction while maintaining computational efficiency. To obtain comprehensive fused features, an Adaptive Feature Fusion (AFF) block merges local and global features at each stage. After the second stage, fused features from the previous stage are connected to the current stage’s AFF block. Inspired by the Adaptive Graph Channel Attention (AGCA) module by Xiang \textit{et al.}~\cite{xiang2023agca}, the AGCA module is placed after both local and global feature extraction stages. It reduces computational redundancy through adaptive graph convolution and enhances feature representation with low computational cost. The computational formula for AGCA is defined as follows:
\begin{equation}
\footnotesize
\begin{aligned}
    &AGCA(\bm m) = \bm m \cdot \text{sigmoid} \left( \bm{F}_r^\prime \left( \text{ReLU} \left( W \cdot \bm {F_r}(GAP(\bm m)) \right. \right. \right. \\
    & \quad \quad \left. \left. \left. \cdot (\bm {A}_0 \times \bm{A}_1 + \bm {A}_2) \right) \right) \right),
\end{aligned}
\end{equation}

\noindent where \(\bm m\) denotes the feature map and $GAP$ represents the Global Average Pooling operation. The functions \( \bm {F}_r \) and \(\bm {F}_r' \) are linear embedding functions, typically implemented using 1×1 convolutions. The \(\bm I \) is the input of the AGCM, and the adaptive adjacency matrix \( \bm A \) comprises three components: \( \bm {A}_0 \) (identity matrix), \( \bm {A}_1 \) (diagonal matrix), and \(  \bm {A}_2 \) (learnable adjacency matrix). The weight matrix \( W \) facilitates the learning relationships between the feature vertices. The formula for computing the fused features is:
\begin{equation*}
\footnotesize
\bm{F^\prime}_{i-1} = \mathrm{AvgPool3D}(\mathrm{Conv3D}(\bm{F}_{i-1})),
\end{equation*}
\begin{equation}
\footnotesize
\label{eq1}
\begin{aligned}
    &\bm {F}_i = \begin{cases}
        \mathrm{GELU}(\mathrm{Conv3D}(\mathrm{LN}(\mathrm{Concat}(\bm{F}_{i-1}', \\
        \quad AGCA(\bm {L}_i), AGCA(\bm {G}_i))))) + \bm{F}_{i-1}'  & \text{if } \bm{F}_{i-1} \neq \text{None} \\
        \mathrm{GELU}(\mathrm{Conv3D}(\mathrm{LN}(\mathrm{Concat}\\(AGCA(\bm L_i), AGCA(\bm G_i)))))  & \text{if } \bm{F}_{i-1} = \text{None},
    \end{cases}
\end{aligned}
\end{equation}
where the \( \bm{L}_{i}\), \( \bm{G}_{i} \), and \( \bm{F}_{i} \) represent the local, global, and fused features at the \( i \)-th stage (\( 1 \leq i \leq 4 \)), respectively.

\subsection{Construction of Hierarchical Graph} \label{S2}
Predicting the malignancy of pulmonary nodules is intricately linked to the complex relationships found within and between multimodal features, as depicted in \textbf{Fig.} \ref{fig2}. To effectively model the morphology of the nodule and fully leverage the relationship between the nodule's growth patterns and clinical risk factors, we designed and constructed a dual-graph structure. This structure comprises intra-modality and inter-modality graphs: the intra-modality graph organizes local features $(L_i)$ and global features $(G_i)$ within the same modality (e.g., CT scan) using fully connected edges, capturing the hierarchical representation of nodule morphology; the inter-modality graph integrates multi-modal features (e.g., fused CT features at different time points, $F_{t0}$, $F_{t1}$, and clinical features, $F_{\text{text}}$) to establish cross-modal associations. All node features are generated by the GLFE and interact through fully connected edges $[E = \{(i,j) \mid i \in \{0, 1, \dots, N-1\}, j \in \{0, 1, \dots, N-1\}, i \neq j\},]$ enabling comprehensive modeling of the multi-modal spatiotemporal evolution characteristics of the nodule.

Let the feature of each node be denoted as \(\mathbf{v}_0, \mathbf v_1, \mathbf\dots,\mathbf v_{n-1} \), where each feature vector has a dimension \( d \). The graph \( \mathcal{G} \) is represented as \( \mathcal{G} = (\mathcal{V}, \mathcal{E}, X) \), where \( \mathcal{V} \) is the set of \( n \) nodes, and \( \mathcal{E} \) is the set of edges, defined as \( \mathcal{E} = \{ (i, j) \mid 0 \leq i, j < n, i \neq j \} \). The node feature matrix \( X \) is given by \( X = [\mathbf v_0~\mathbf v_1~ \mathbf \dots~\mathbf v_{n-1}]^T \), with \( t_i \) being the feature vector of the \( i \)-th node. Finally, we use a GAT to capture interactions within the modality graph.

\begin{table*}[t]
\tiny

\resizebox{\textwidth}{0.08\textheight}{
\begin{tabular}{cc|ccccc|ccccc}
\bottomrule
\multicolumn{1}{l}{} & \multicolumn{1}{l|}{} & \multicolumn{5}{c|}{ {\textbf{NLST-cmst~\cite{kramer2011lung}}}}                                                         & \multicolumn{5}{c}{ {\textbf{CLST~\cite{jian2024cross}}}}  \\
\hline
 \textbf{Method}      & Param$(\downarrow)$  & Acc$\uparrow$ &  {Pre$(\uparrow)$} &  {F1$(\uparrow)$} & AUC$(\uparrow)$    &  {Rec$(\uparrow)$} &  {Acc$(\uparrow)$} &  {Pre$(\uparrow)$} &  {F1$(\uparrow)$} &  {AUC$(\uparrow)$}    &  {Rec$(\uparrow)$} \\
 \hline
Scans~\cite{veasey2020lung}         & 8.10M                & 84.62            & 79.31             & 79.31            & 86.29          & 79.31          & 51.43            & 72.22             & 60.47            & 55.60          & 52.00          \\
DeepCAD~\cite{aslani2024enhancing}      & 11.4M               & 85.18            & 84.00             & 81.48            & 87.13          & 75.86          & 57.80            & 76.20             & 68.00            & 62.50          & 60.00          \\
NAS-Lung~\cite{jiang2021learning}      & 57.3M               & 82.05            & 78.57             & 75.86            & 80.35          & 73.33          & 47.14            & 60.00             & 45.00            & 47.20          & 52.00          \\
ICHPro~\cite{yu2024ichpro}        & 58.2M               & 83.99            & 82.79             & 78.71            & 89.61          & 76.56          & 60.71            & 75.20             & 68.57            & \textcolor{blue}{67.50}          & 60.00          \\
CSF-Net~\cite{10981213}       & 10.1M              & 86.52            & 84.85             & 82.35            & 90.53          & 81.71          & 61.51            & 72.22             & 71.00            & 62.10          & 60.00          \\
MMFusion~\cite{wu2024mmfusion}       & 15.3M              & 85.16            & 86.00             & 83.02            & 90.89          & \textcolor{blue}{87.57}          & 61.51            & 75.20             & 72.80            & 63.30          & 68.00          \\
HGCN~\cite{jiang2024hyperrep}          & \textcolor{blue}{5.86M}               & \textcolor{blue}{87.66}            & \textcolor{blue}{87.14}             & 83.90            & \textcolor{blue}{92.22}          & 77.34          & \textcolor{blue}{62.86}            & 75.00             & \textcolor{blue}{73.47}            & 57.60          & \textcolor{blue}{72.00}          \\
SAG-VIT~\cite{venkatraman2025sagvitscaleawarehighfidelitypatching}       & 6.69M               & 87.02            & 84.16             & \textcolor{blue}{84.25}            & 91.21          & 82.92          & \textcolor{blue}{62.86}            & \textcolor{blue}{77.78}             & 61.18            & 62.00          & 58.00          \\
\hline
\textbf{DGSAN (Our)}        & \textbf{\textcolor{red}{4.21M}}      & \textbf{\textcolor{red}{90.79}}   & \textbf{\textcolor{red}{88.87}}    & \textbf{\textcolor{red}{88.14}}   & \textbf{\textcolor{red}{93.92}} & \textbf{\textcolor{red}{92.86}} & \textbf{\textcolor{red}{65.28}}   & \textbf{\textcolor{red}{79.31}}    & \textbf{\textcolor{red}{77.20}}   & \textbf{\textcolor{red}{69.80}} & \textbf{\textcolor{red}{78.00}}\\
\toprule
\end{tabular}
}
\caption{Comparison of methods on the NLST-cmst dataset and CLST dataset (\%). The best and second-best results are shown in \textcolor{red}{red} and \textcolor{blue}{blue}, respectively.}
\label{tab1}
\end{table*}

\begin{figure*}[]
    \centering
    \includegraphics[width=0.85\textwidth]{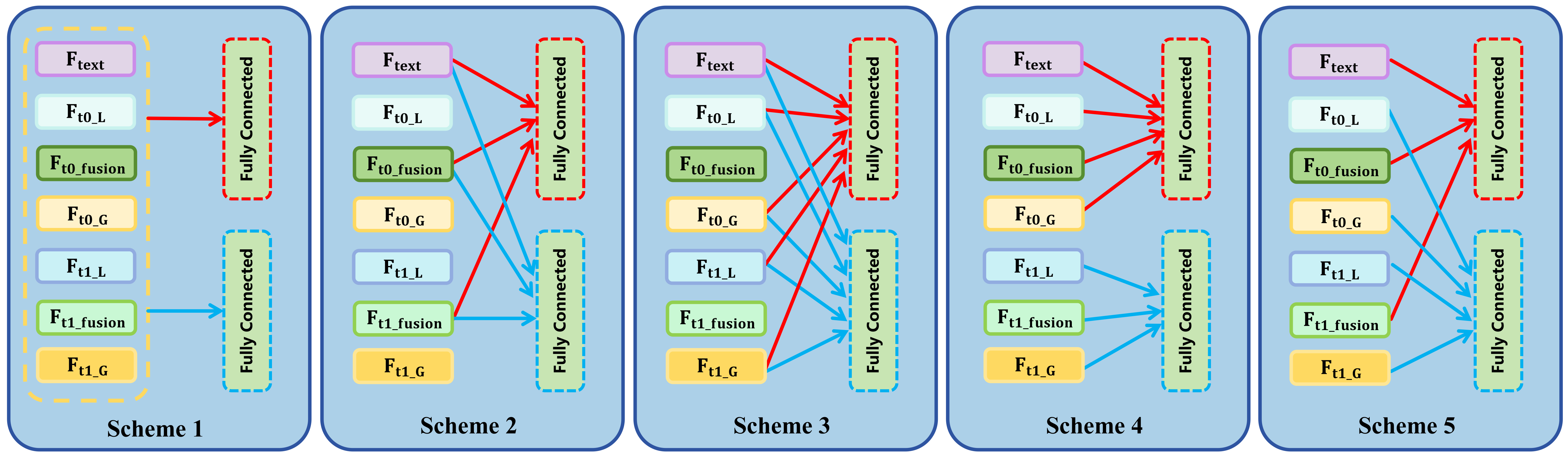}
    \caption{Five modal graph construction schemes. From left to right: Separate graphs for each modality, No local/global features used, No feature fusion, Separate graphs for relevant features at two-time points, and Our custom scheme.    
}
    \label{fig3}
\end{figure*}
\subsection{Hierarchical Cross-Modal Graph Fusion Module} \label{S3}
The HCMGFM comprises two Self-Attention Blocks (SAB) and one Cross-Attention Block (CAB). The initial layer features a dual-path parallel SAB, which independently processes dual-modal input features. This layer utilizes an adaptive weight allocation mechanism to focus on higher-order semantic relationships within each modality, thereby enhancing and refining features specific to each modality. The CAB in the middle establishes bidirectional information pathways between the modalities. It uses the key-value query mechanism of Multi-head Cross-Attention to create a modality association matrix, facilitating the interaction between different modalities. In the final layer, another SAB is employed to perform a global recalibration of the fused joint features. This recalibration optimizes attention weights in the fully connected feature space, reducing potential alignment noise between modalities and enhancing both semantic consistency and spatial coherence of the fused features.

To better achieve progressive fusion of intra-modality and inter-modality feature information, we have designed a hierarchical structure consisting of "self-attention $\rightarrow$ cross-attention $\rightarrow$ self-attention." This design complements our dual-graph construction approach, effectively capturing spatiotemporal features and multi-modal dependencies, while overcoming the limitations of simple fusion methods.

\section{Experiments and Results}
\subsection{Datasets and Implementation Details}
\textbf{The NLST-cmst Dataset~\cite{kramer2011lung}}. The NLST-cmst dataset originates from the National Lung Screening Trial (NLST), conducted by the National Cancer Institute (NCI). It comprises data from 433 subjects, each with 3D Regions of Interest (RoIs) for lung nodules sized 16×64×64, which have been pathologically classified as either benign or malignant. Participants underwent at least two longitudinal CT scans, with additional clinical information, such as age, gender, smoking history, and screening outcomes, collected concurrently. Obtaining multi-center, multi-timepoint longitudinal imaging required consistent subject participation over a lengthy period for multiple radiation scans, alongside rigorous standardization of scanning protocols and equipment settings. Furthermore, achieving precise RoI annotation depended heavily on extensive quality control carried out by experienced imaging specialists.

Given these technical and resource-intensive demands, datasets that integrate systematic follow-up imaging with comprehensive clinical annotations for lung nodules are exceedingly rare. The NLST-cmst dataset is divided into a training set of 347 cases and a test set of 86 cases, maintaining a 4:1 ratio. This configuration provides a valuable and reliable resource for investigating the dynamic evolution prediction of lung nodules.

\begin{table}[t]
    \centering
    
    \resizebox{\linewidth}{!}{ 
        \begin{tabular}{cccccc}
            \toprule
            \textbf{Method}             &  {Acc$(\uparrow)$} &  {Pre$(\uparrow)$} &  {F1$(\uparrow)$} &  {AUC$(\uparrow)$} &  {Rec$(\uparrow)$} \\
            \midrule
            t0 image                    & 85.53            & 77.42             & 81.36            & 85.25          & 82.71          \\
            t0 image+clinical           & 86.84            & 76.47             & 83.87            & 87.59          & \textcolor{blue}{86.86} \\
            t1 image                    & 87.24            & 80.27             & 81.47            & 87.59          & 84.86          \\
            t1 image+clinical           & 88.16            & 81.33             & 82.35            & 88.54          & 83.34          \\
            without GFF                 & \textcolor{blue}{88.47} & \textcolor{blue}{85.46} & \textcolor{blue}{85.19} & \textcolor{blue}{91.47} & 82.14 \\
            without HCC+GFF            & 86.84            & 80.13             & 83.35            & 86.43          & 85.34          \\
            \textbf{DGSAN (Our)}        & \textbf{\textcolor{red}{90.79}} & \textbf{\textcolor{red}{88.87}} & \textbf{\textcolor{red}{88.14}} & \textbf{\textcolor{red}{93.82}} & \textbf{\textcolor{red}{92.86}} \\
            \bottomrule
        \end{tabular}
    }
    \caption{Ablation study of the proposed method on the NLST-cmst dataset (\%). The best and second-best results are shown in \textcolor{red}{red} and \textcolor{blue}{blue}, respectively.}
    \label{table2}
\end{table}

\begin{figure}[t]
    \centering
    \includegraphics[width=0.8\linewidth]{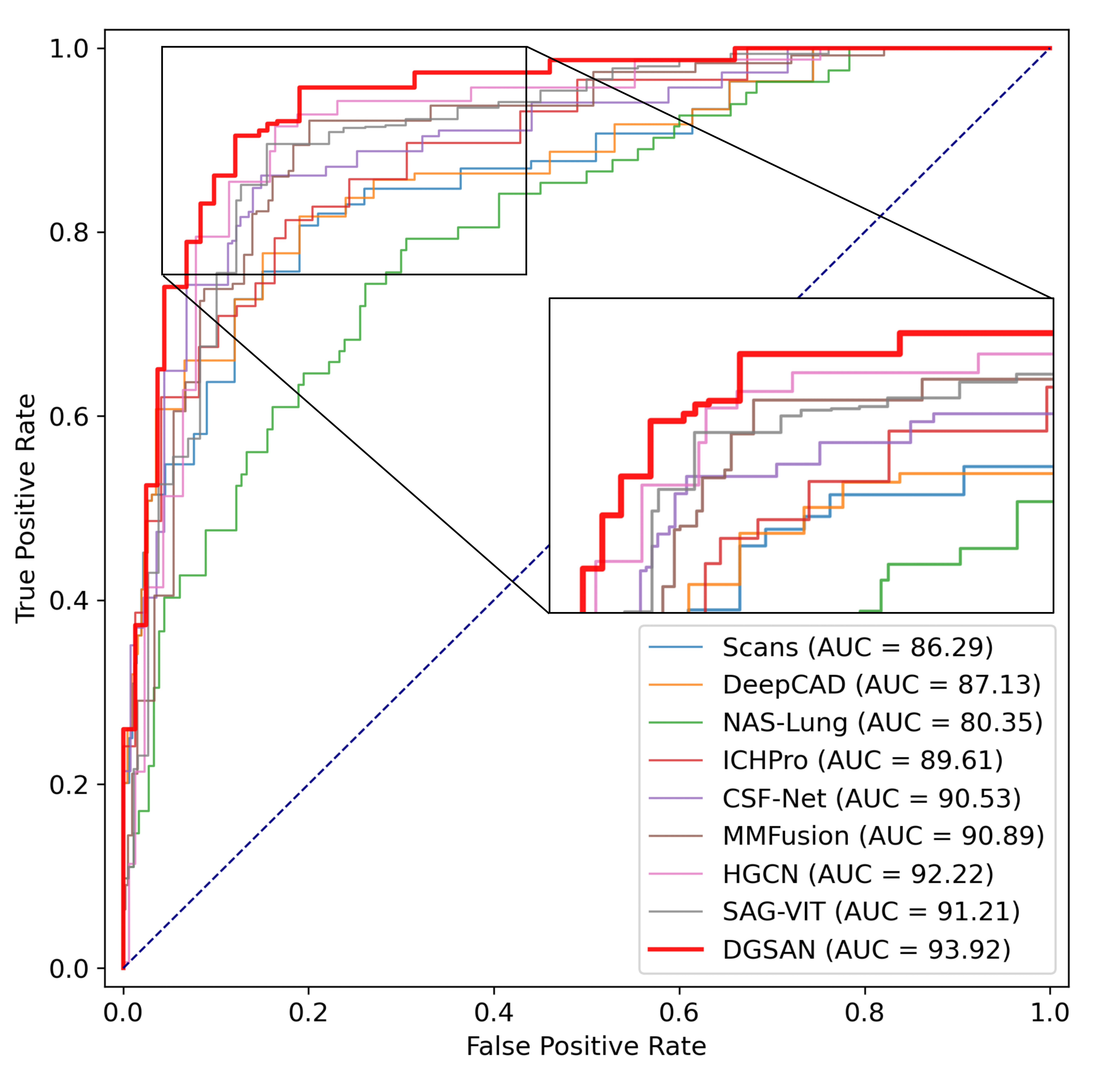} 
    \caption{ROC curves comparison between DGSAN and other methods on the NLST-cmst dataset.}
    \label{fig4}

\end{figure}

\textbf{The CLST Dataset~\cite{jian2024cross}}. 
The CLST dataset encompasses 109 patients, capturing 317 CT sequences and detailing 2,295 annotated nodules. These nodules are classified as malignant, including invasive adenocarcinoma, microinvasive adenocarcinoma, in-situ adenocarcinoma, and other malignant types, and benign, which includes inflammation and other benign categories. For external validation purposes, we selected 36 cases, each with two time points, resulting in a total of 72 data points. This set was exclusively used for testing and did not contribute to model training. From these datasets, 3D RoIs measuring 16×64×64 were derived based on nodule locations, with nodule diameters recorded as clinical features.

Despite the dataset only providing diameter information, clinical diagnoses generally consider a mix of texture characteristics, dynamic growth patterns, and the complex interface between the nodule and lung tissue. Therefore, in our model, diameter serves as a supplementary input rather than the primary determinant for classifying malignancy. This approach ensures a more holistic understanding of the nodule characteristics.

\textbf{The Implementation Details}. We implemented our method using the torch-2.1.0-cu12.1-cudnn8.9 framework on a GeForce RTX 3090Ti GPU. We first pre-trained the GLFE with cross-entropy loss, the Adam optimizer, a learning rate of 0.0001, and 200 epochs. Then, we trained the entire DGSAN model under the same settings. Momentum parameters were set as $\beta$1 = 0.5 and $\beta$2 = 0.999, with parameter updates every 20 epochs. Given the dataset limitations, we used 5-fold cross-validation to evaluate the model's performance, ensuring the reliability and generalizability of the results.

\subsection{Main Results}
\textbf{Comparison Experiment.}
In this experiment, we compared our proposed method with eight other models: Scans~\cite{veasey2020lung}, DeepCAD~\cite{aslani2024enhancing}, NAS-Lung~\cite{jiang2021learning}, ICHPro~\cite{yu2024ichpro}, CSF-Net~\cite{10981213}, MMFusion~\cite{wu2024mmfusion}, HGCN~\cite{jiang2024hyperrep}, and SAG-VIT~\cite{venkatraman2025sagvitscaleawarehighfidelitypatching}. Each model offers a unique approach for nodule classification or other tasks, of which MMFusion, HGCN, and SAG-VIT are graph-based methods.

\begin{table}[t]
    \centering
    \setlength{\tabcolsep}{1pt}
    
    \tiny 
    \resizebox{\linewidth}{0.065\textheight}{
    \begin{tabular}{cccccc}
        \toprule
        \textbf{Method} &  {Acc$ ( \uparrow )$} &  {Pre$(\uparrow)$} &  {F1$(\uparrow)$} &  {AUC$(\uparrow)$} &  {Rec$(\uparrow)$} \\
        \midrule
        Scheme 1          & 88.21          & 86.91          & 86.82          & 90.49          & 90.10          \\
        Scheme 2          & 88.01          & 86.12          & 87.74          & 91.80          & 91.22          \\
        Scheme 3          & 88.93          &  \textcolor{blue}{87.80}          & 87.38          & 91.17          &  \textcolor{blue}{91.54}          \\
        Scheme 4          &  \textcolor{blue}{90.27}          & 87.51          &  \textcolor{blue}{87.93}          &  \textcolor{blue}{92.17}          & 91.32          \\
        \textbf{Scheme 5 (Our)} & \textbf{\textcolor{red}{90.79}} & \textbf{\textcolor{red}{88.87}} & \textbf{\textcolor{red}{88.14}} & \textbf{\textcolor{red}{93.82}} & \textbf{\textcolor{red}{92.86}} \\
        \bottomrule
    \end{tabular}
    }
    \caption{Modal graph construction scheme study on NLST-cmst dataset (\%). The best and second-best results are shown in \textcolor{red}{red} and \textcolor{blue}{blue}, respectively.}
    \label{table3}
\end{table}

\begin{table}[t]
    \centering
    \setlength{\tabcolsep}{1pt}
    
    \resizebox{\linewidth}{!}{ 
    \begin{tabular}{cccccc}
        \toprule
        \textbf{Method} &  {Acc$(\uparrow)$} &  {Pre$(\uparrow)$} &  {F1$(\uparrow)$} &  {AUC$(\uparrow)$} &  {Rec$(\uparrow)$} \\
        \midrule
        CAB-CAB              & 89.83            & 87.92             & 87.89            & 92.84          & 90.16          \\
        SAB-SAB              & 89.13            & 88.09             & 87.67            & 92.23          & 91.79          \\
        SAB-CAB              & 90.19            & 88.32             & 87.89            & 92.52          & 92.23          \\
        CAB-SAB              & \textcolor{blue}{90.43} & 88.53 & \textcolor{blue}{88.01} & \textcolor{blue}{93.59} & 91.98 \\
        CAB-SAB-CAB          & 89.53            & \textcolor{blue}{88.74} & 87.72            & 92.68          & \textcolor{blue}{92.64} \\
        \textbf{SAB-CAB-SAB (Our)} & \textbf{\textcolor{red}{90.79}} & \textbf{\textcolor{red}{88.87}} & \textbf{\textcolor{red}{88.14}} & \textbf{\textcolor{red}{93.82}} & \textbf{\textcolor{red}{92.86}} \\
        \bottomrule
    \end{tabular}
    }
    \caption{Structural design study of the HCMGFM on NLST-cmst dataset (\%). The best and second-best results are shown in \textcolor{red}{red} and \textcolor{blue}{blue}, respectively.}
    \label{table4}
\end{table}

To ensure a fair comparison, we maintained consistent training environments and hyperparameter settings across all models. We evaluated our proposed model against these eight models using six key metrics: Parameter Count (Param), Accuracy (Acc), Precision (Pre), F1 score (F1), Area Under the Curve (AUC), and Recall (Rec). These metrics provide a comprehensive assessment of predictive accuracy, precision, and overall performance. As shown in \textbf{Table} \ref{tab1} and \textbf{Fig.} \ref{fig4}, our proposed model outperformed all the other models across all metrics. Furthermore, the DGSAN demonstrates outstanding performance despite a reduction in parameters by at least 28.16\%. On the NLST-cmst dataset, accuracy improved by at least 3.57\%, while the untrained CLST dataset saw an improvement of at least 3.85\%. These findings suggest that DGSAN is not only efficient and lightweight but also excels in nodule classification.

\textbf{Ablation Study.}
To evaluate the model's performance in combining multimodal data and its modular design, we performed a series of ablation experiments. Initially, we used only the data from missing modalities. Subsequently, we incrementally removed the Dual-Modal Graph Construction (HCC) and Graph Feature Fusion (GFF) components. As illustrated in Table \ref{table2}, the integration of multimodal data resulted in improvements in accuracy, precision, F1 score, AUC, and recall by at least 2.62\%, 3.99\%, 3.46\%, 2.57\%, and 6.91\%, respectively. These improvements underscore the benefits of utilizing cross-modal features. On the other hand, the removal of any submodule caused a decline in these performance metrics, emphasizing the critical roles of HCC and the fusion of dual-modal graph features.


\textbf{Modal Graph Construction Scheme.}
To validate the superiority of our proposed inter- and intra-modal graph construction scheme, we experimented with various combinations involving seven different modality features and five distinct graph construction schemes, as illustrated in \textbf{Fig.} \ref{fig3}. The results, detailed in \textbf{Table} \ref{table3}, compared to other schemes, the Acc, Pre, F1, AUC, and Rec metrics of our proposed method have increased by at least 0.58\%, 1.22\%, 0.24\%, 1.79\%, and 1.44\%, respectively. This finding strongly supports the rationality and effectiveness of our graph construction approach for modality data.

\textbf{Hierarchical Cross-Modal Graph Fusion Module.} We further investigated various structural designs by arranging the CAB and SAB in different configurations to determine the optimal arrangement. As shown in \textbf{Table} \ref{table4}, the Acc, Pre, F1, AUC, and Rec metrics of the HCMGFM have increased by at least 0.40\%, 0.15\%, 0.15\%, 0.25\%, and 0.24\%, respectively. The results show that the three-stage structure effectively fuses bimodal semantics, significantly enhancing the model's performance.

\section{Conclusion}
We introduced the DGSAN, a novel graph-based classification model for predicting key characteristics of pulmonary nodules. DGSAN leverages an GLFE module to capture multi-scale features from multitemporal nodule data, constructs a dual-modality graph to model complex relationships and higher-order dependencies, and employs an HCMGF module for a deep semantic fusion of multimodal information. Experimental results show that DGSAN outperforms existing methods in predicting pulmonary nodule malignancy while maintaining high computational efficiency. Future work will focus on evaluating the model with larger datasets and exploring advanced fusion techniques to further improve the generalization and predictive performance of pulmonary nodules diagnosis.

\section{Acknowledgments}
This work was supported by Guangxi Key R\&D Project  [No. AB24010167]; Guangxi Science and Technology Program [No. FN2504240022]; Shenzhen Medical Research Fund [No. C2401036, 20232ABC03A25]; Guangdong Basic and Applied Basic Research Foundation [Nos. 2025A1515011617, 2022A1515110570]; National Natural Science Foundation of China [No. 62076084, 61702146, U20A20386, U22A2033]; Zhejiang Provincial Natural Science Foundation of China [No. LY21F020017, 2023C03090]; Open Project Program of the State Key Laboratory
of CAD\&CG [No. A2410], Zhejiang University; Shenzhen Science and Technology Innovation Program [No. CJGJZD20220517142000002]; Shenzhen Longgang District Science and Technology Innovation Special Fund [No. LGKCYLWS2023018]; and Central Funds Guiding the Local Science and Technology Development Project [No. 2025ZYDF106].
\bibliography{CR/9448}

\end{document}